\documentclass{article}
\usepackage[utf8]{inputenc}
\usepackage{url}
\usepackage{hyperref}
\usepackage{authblk}
\usepackage[nottoc]{tocbibind}
\usepackage{amsmath}
\usepackage{graphicx} 
\usepackage{amsfonts}

\usepackage{booktabs}
\usepackage{multirow}
\usepackage{adjustbox}

\usepackage[skip=5pt]{caption} 

\newtheorem{theorem}{Theorem}
\title{New Statistical Framework for Extreme Error Probability in High-Stakes Domains for Reliable Machine Learning}
\author[1,2]{Umberto Michelucci}
\author[3,2]{Francesca Venturini}

\affil[1]{Computer science Department, Lucerne University of Applied Sciences and Arts, Luzern 6002, Switzerland}
\affil[2]{TOELT LLC, Machine Learning Research and Development Department, Winterthur 8406, Zurich, Switzerland }

\affil[3]{Institute of Applied Mathematics and Physics, ZHAW - Zurich University of Applied Sciences, Winterthur 8400, Zurich, Switzerland}

\begin{document}

\maketitle


\begin{abstract}
Machine learning is vital in high-stakes domains, yet conventional validation methods rely on averaging metrics like mean squared error (MSE) or mean absolute error (MAE), which fail to quantify extreme errors. Worst-case prediction failures can have substantial consequences, but current frameworks lack statistical foundations for assessing their probability.
In this work a new statistical framework, based on Extreme Value Theory (EVT), is presented that provides a rigorous approach to estimating worst-case failures. 
Applying EVT to synthetic and real-world datasets, this method is shown to enable robust estimation of catastrophic failure probabilities, overcoming the fundamental limitations of standard cross-validation. This work establishes EVT as a fundamental tool for assessing model reliability, ensuring safer AI deployment in new technologies where uncertainty quantification is central to decision-making or scientific analysis.
\end{abstract}

\section{Introduction}

Machine learning models are increasingly used in high-stakes applications, from healthcare diagnostics to autonomous systems, where understanding and quantifying model uncertainty is vital. The ability to ensure that models generalise well to unseen data is a central challenge in modern AI research, especially as models become more complex and computationally intensive. As new technologies push the boundaries of AI-driven discovery, the need to assess worst-case errors has become fundamental, particularly in domains where incorrect predictions can lead to severe consequences \cite{kompa_second_2021, chua_tackling_2023, seoni_application_2023, tang_prediction-uncertainty-aware_2022, zhu_know_2020}. Despite the widespread adoption of machine learning validation techniques, risk assessment of extreme prediction errors remains impossible, limiting the reliability, trustworthiness, and applicability of AI systems.

A limitation of existing model validation methods is the focus on averaged performance metrics, such as mean squared error (MSE) or mean absolute error (MAE), which fail to capture extreme errors. Standard approaches, including cross-validation (CV), provide insight into generalisation but do not explicitly quantify worst-case risks \cite{michelucci2024fundamental}. However, in real-world applications, the largest error that a model can produce is often of greatest concern. For example, in medicine, an underestimated disease progression may delay critical treatments, and in financial forecasting, unaccounted extreme deviations may lead to substantial economic loss. The failure to rigorously assess extreme errors results in limiting the practical adoption of machine learning in ubiquitous real-world scenarios.
Despite the use of advanced cross-validation techniques, including Monte Carlo and $k$-fold CV, existing methods remain inadequate in quantifying tail-end behaviour of model errors. 

In this work, we propose the application of Extreme Value Theory (EVT), a well-established statistical framework for modelling rare events. 
Specifically, this work has three major contributions. First, we present a novel statistical framework designed to integrate seamlessly with Monte-Carlo cross validation techniques. This new framework can estimate the statistical characteristics of the maximum error expected in the performance of a machine learning model. This methodology addresses the fundamental limitation of traditional validation metrics, such as mean squared error (MSE), which fail to quantify worst-case errors. Second, the application of this new approach is shown on synthetic data and on two real dataset, to showcase its ability to assess extreme errors in predictions. Third, this work describes in detail the algorithm for the practical implementation of this approach.

\section{Results}

\label{sec:exp}

In this section, we provide experimental evidence to validate the proposed methodology. The experiments are designed to demonstrate how the proposed EVT-based approach can be used to quantify and predict extreme errors in regression tasks. To this end, we first used synthetic data to illustrate the process in a controlled environment in Section \ref{sec:synthgen} and then validated the approach on real-world datasets in Section \ref{sec:questionc}. 

For a more comprehensive discussion of EVT, including its theoretical foundations and practical implementation, the reader is referred to Section \ref{sec:methods}. However, before presenting our results, it is useful to provide an intuitive overview of EVT.
Extreme Value Theory (EVT) is a statistical framework designed to model and analyse rare, extreme events by focussing on the tails of probability distributions of extremes. Unlike classical statistical approaches that primarily describe the central tendencies of data (usually averages), EVT provides tools for quantifying the probability and magnitude of extreme deviations (the tail behaviour of error distributions). It offers asymptotic results for the distribution of maxima (or minima) of a sequence of random variables, leading to two fundamental distribution families: the Generalized Extreme Value (GEV)  and the Generalised Pareto Distribution (GPD) distributions.

The GEV distribution characterises the limiting behaviour of block's maxima, where data are partitioned into blocks (e.g., yearly or monthly maxima), and the largest value in each block is considered. In contrast, the GPD  models exceedances over a predefined threshold, allowing for a more flexible and detailed assessment of tail behaviour. The algorithm for adapting this approach for use in machine learning applications (the main contribution of this work) is presented in Section \ref{sec:evtml}..

\subsection{EVT Applied to Synthetic Data}
\label{sec:synthgen}
\begin{figure}
    \centering
    \includegraphics[width=0.9\linewidth]{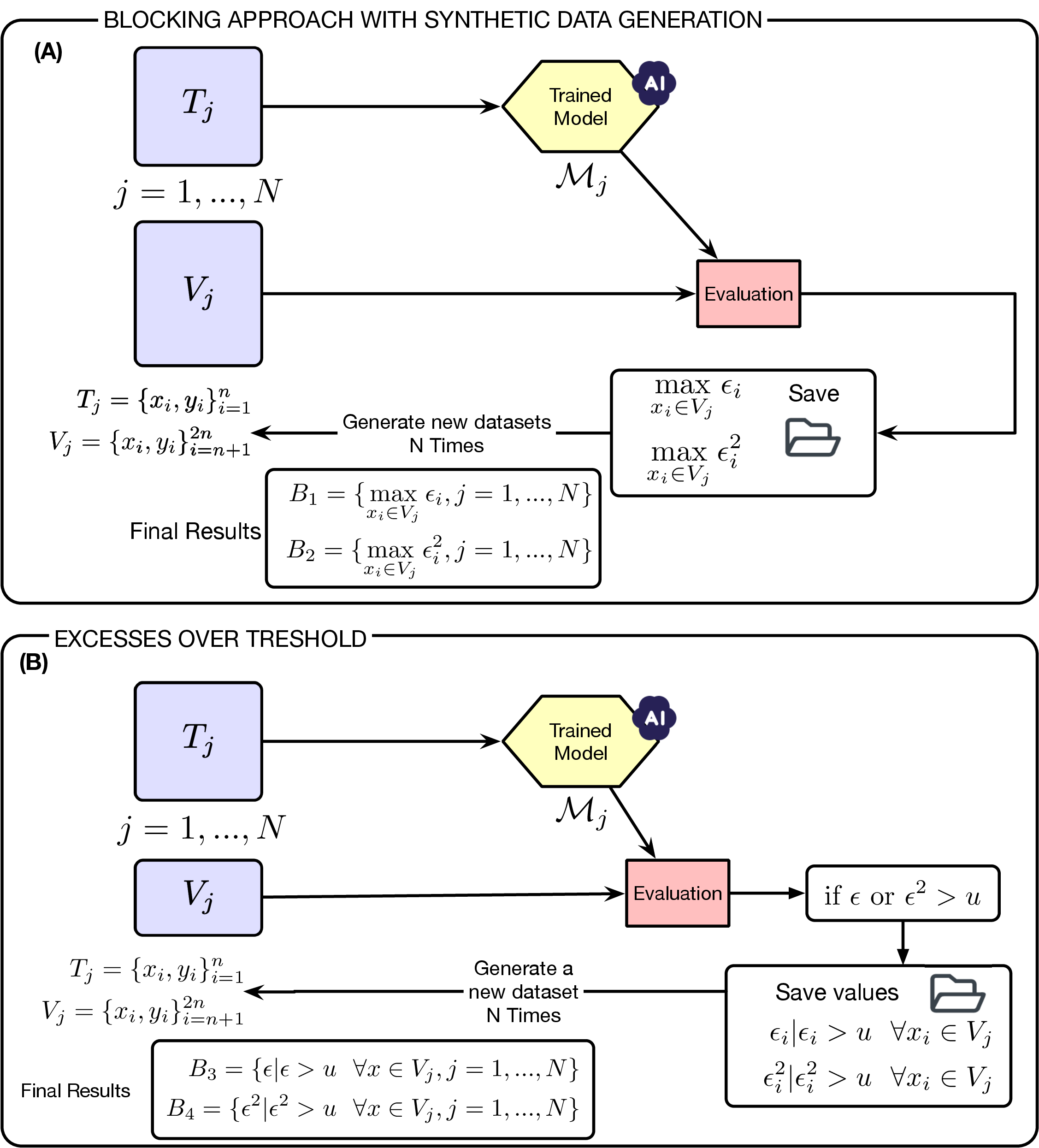}
    \caption{Process workflow for applying EVT to machine learning with synthetic data. Panel (A) shows the blocking method, while Panel (B) illustrates the threshold-based approach. }
    \label{fig:blocking}
\end{figure}

To introduce the application of EVT to machine learning, let us consider first a synthetically generated dataset. The data consists of tuples of real numbers $( x_i, y_i)$ with  $x_i\in [-5,5]$ and $y$ given by
\begin{equation}
    y_i = x_i^2+\epsilon
\end{equation}
with $\epsilon$ a random number number between -5 and 5 sampled from a uniform distribution.
$2n$ tuples will be built this way to form two datasets that will be indicated with $T_j= \{ x_i, y_i\}_{i=1}^n$ and $V_j=\{ x_i, y_i\}_{i=n+1}^{2n}$. This process is repeated $N$ times, so that $N$ datasets $T_j$ and $V_j$ will be created (with $j=1,\cdots, N$). With each of these $N$ datasets $T_j$ a model ${\cal M}_j$ will be trained and it will be used to predict $y_i$ from the validation datasets $V_j$. These predictions will be indicated by ${\cal M}_j(x_i)$. Square and absolute errors for each observations  are given by 
\begin{equation}
    \epsilon_i^2 = ({\cal M}_j(x_i)-y_i)^2
\end{equation}
and 
\begin{equation}
    \epsilon_i=|{\cal M}_j(x_i)-y_i|
\end{equation}
respectively (for the dataset $V_j$). The following quantities can be evaluated. 
\begin{equation}
    G^j_n=\max_{x_i \in V_j} \epsilon_i 
\end{equation}
and 
\begin{equation}
    M_n^j=\max_{x_i \in V_j} \epsilon_i^2 
\end{equation}
The process is depicted in detail in panel (A) of Figure \ref{fig:blocking} . At the end of the process, two datasets are generated: $B_1 = \{ G^j_n | \ \forall \ V_j, j=1,...,N\}$ and $B_2 = \{M_n^j |  \ \forall \  V_j, j=1,...,N\}$. These will be used in the blocking approach.

Furthermore, it is possible, defining $u>0, u \in \mathbb{R}$, to study the distribution of the values in the sets $B_3  = \{\epsilon_i | \epsilon_i >u  \ \ \forall x_i \in V_j\ \forall j=1,...,N\}$ and $B_4  = \{\epsilon^2_i | \epsilon_i^2 >u  \ \ \forall x_i \in V_j \ \forall j=1,...,N\}$.
This process is depicted in panel (B) in Figure \ref{fig:blocking} and will be used in the excess over a treshold approach.
The confidence intervals for the parameters are evaluated with a conventional bootstrap approach.

\subsubsection{Distribution of Extreme Errors with Synthetic Data with the Blocking Approach}
\label{sec:questiona}

\begin{figure}[hbt]
    \centering
  \includegraphics[width=1\linewidth]{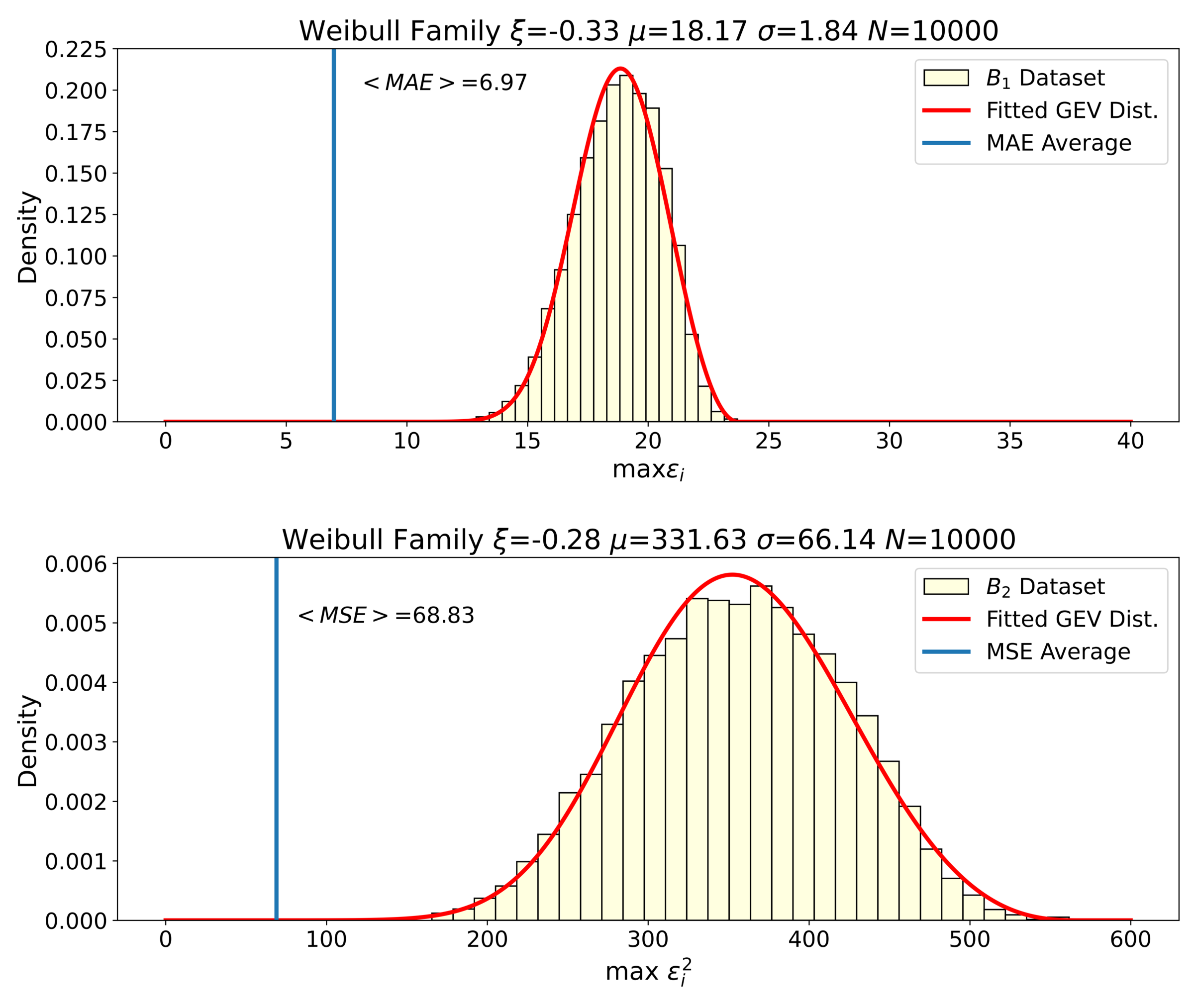}
    \caption{Distribution of $G^j_n$ (in panel (A)) and $M_n^j$ (in panel (B)). The blue vertical line indicates the average of the metric MAE in panel (A) and MSE in panel (B). It is important to clarify that although the symbol $\mu$ is commonly associated with averages, in this context, it does not represent the mean of the values.}
    \label{fig:results_a_1}
\end{figure}

Let us consider $N=10^4$ values of $M_n^j$ and $G^j_n$ generated according to the procedure described in Section \ref{sec:synthgen} with $n=50$ and $\cal{M}$ chosen as linear regression (comparison of multiple models can be found in Subsection \ref{sec:comparison}). The resulting distributions of the $G^j_n$ and $M_n^j$ are shown in Figure \ref{fig:results_a_1}, Panel A and B respectively (for details on the meaning of the fitted parameters, the reader is referred to the Methods section \ref{sec:methods}). 

By calculating the 95\% quantile of the fitted GEV distributions of $G^j_n$ (Panel (A) in Figure \ref{fig:results_a_1}), it is now possible to calculate that linear regression will give an error not larger than 27.4 in 95\% of the cases. One can reformulate this by saying that with a confidence of 95\%, linear regression will give an error smaller than 27.4. This value is, as is evident, much higher than $<\text{MAE}>=6.97$ normally used in classical model validation approaches. The key difference here is that while the MAE provides an average measure of error, it does not capture the probability of extreme deviations, which is crucial in risk-sensitive applications. One can also use this method, for example, to determine that linear regression will always give an error smaller than, say, 18.0 in 35\% of cases. This highlights the probabilistic nature of error distribution: rather than relying on a single average metric, EVT allows us to quantify the likelihood of different error magnitudes. This perspective is particularly useful in scenarios where occasional large errors can have significant consequences, such as financial forecasting or safety-critical systems. For $M_n^j$ it can be equivalently said that  linear regression will give a squared error not larger than 465.7 in 95\% of the cases.
In both cases, the fitted distributions are of the Weibull family ($\xi<0$). The blue vertical line in Figure \ref{fig:results_a_1} indicates the MAE (panel A) and MSE (panel B) respectively. For example, in panel (A) of Figure \ref{fig:results_a_1} it is reporeted how $<\text{MAE}>=6.97$, a much lower value of the most probable largest error (around 18). The Figure illustrates that judging a model performance by the average of a metric is at best superficial, and at worst deceptive and might lead to wrong conclusions. 

\begin{figure}[hbt]
    \centering
    \includegraphics[width=1\linewidth]{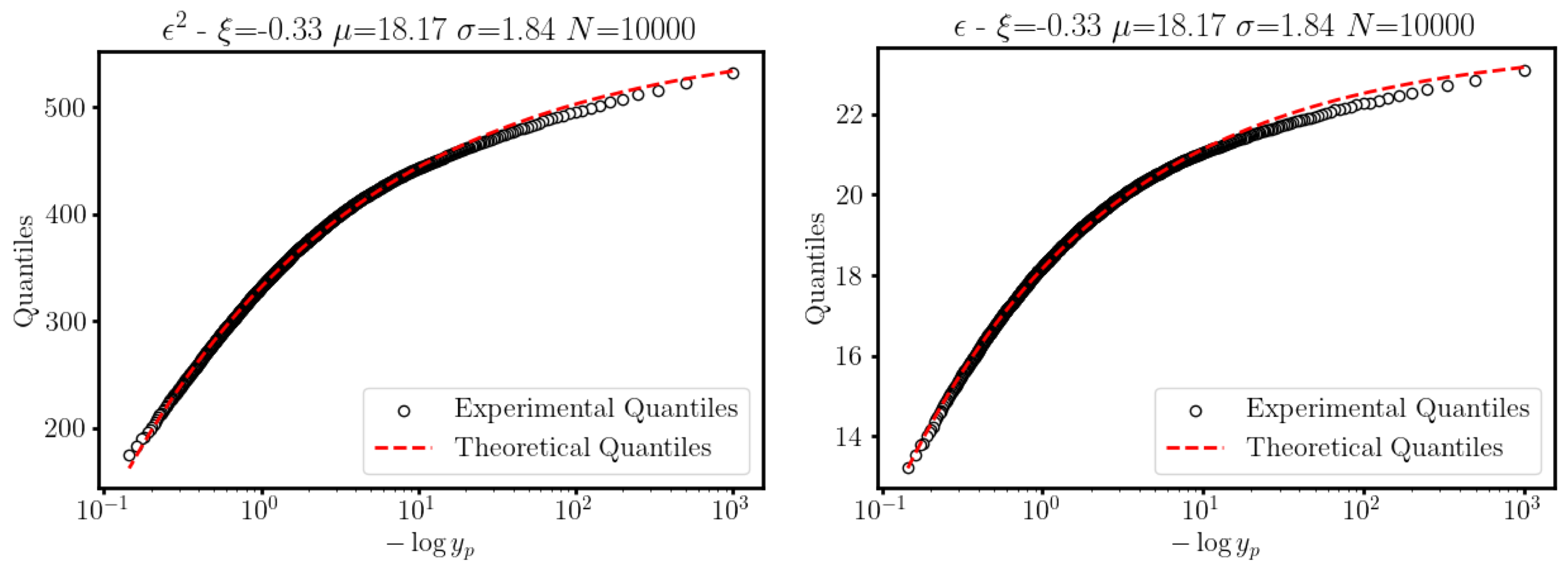}
    \caption{Return plots to assess the goodness of the fits of the data to the GEV distribution families.}
    \label{fig:returnplots1}
\end{figure}

For completeness, in Figure \ref{fig:returnplots1}, one can see the return level plots \cite{coles_introduction_2001} for the distributions of $G^j_n$ and $M_n^j$. 
The plots should be intepreted according to the following two main points.
\begin{itemize}
    \item $\xi=-0.33$: Indicates a Weibull-type tail, meaning the errors have an upper bound. This implies that the extreme $\epsilon_i$ or $\epsilon_i^2$ values will not grow indefinitely.
    \item The fit between experimental and theoretical quantiles is strong, suggesting that EVT accurately models the extremes of the $\epsilon_i$ and $\epsilon_i^2$  distributions.
\end{itemize}

In general it is clear that there is a strong agreement between the theoretical and experimental quantiles, which suggests that the parameters derived from the fitting process are accurate.

\subsubsection{Distribution of  Excesses over a Treshold with Synthetic Data}
\label{sec:questionb}

\begin{figure}[hbt]
    \centering
    \includegraphics[width=1\linewidth]{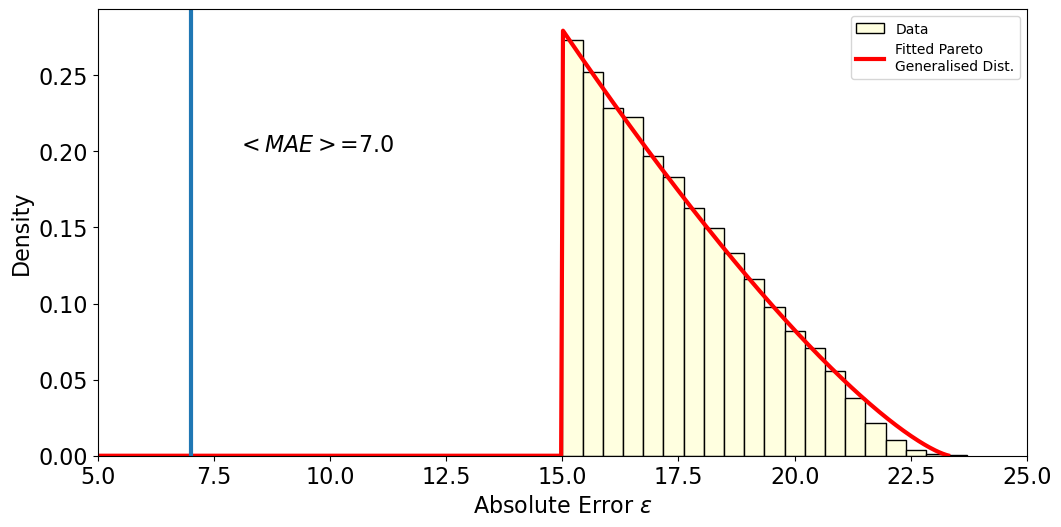}
    \caption{The analysis conducted using datasets $B_3$ and $B_4$, as outlined in the text, reveals the distribution of $\epsilon$ values exceeding the threshold of 15, depicted as light yellow bars. The red line represents the fitted generalized Pareto distribution, characterized by parameters $\xi=-0.43$, $u=15$, and $\sigma=3.57$.}
    \label{fig:pareto1}
\end{figure}

The datasets $B_3$ and $B_4$ can similarly undergo an analysis as previously outlined. Figure \ref{fig:pareto1} illustrates the distribution of $\epsilon$ values that surpass a threshold of 15. The red line represents the fitted generalized Pareto distribution, characterized by parameters $\xi = -0.43$, $u = 15$, and $\sigma = 3.57$. For conciseness, only the plot for $\epsilon$ is presented, as the distribution for $\epsilon^2$ would be analogous and does not contribute further to the discussion. Also in this case, one can now calculate that linear regression will do an error not larger than 21.0 in 95\% of the cases. Looking at it from a different angle, linear regression will give an error smaller than, say, 19 in 79\% of predictions.

\subsubsection{Comparison of Multiple Models}
\label{sec:comparison}

\begin{figure}
    \centering
    \includegraphics[width=1\linewidth]{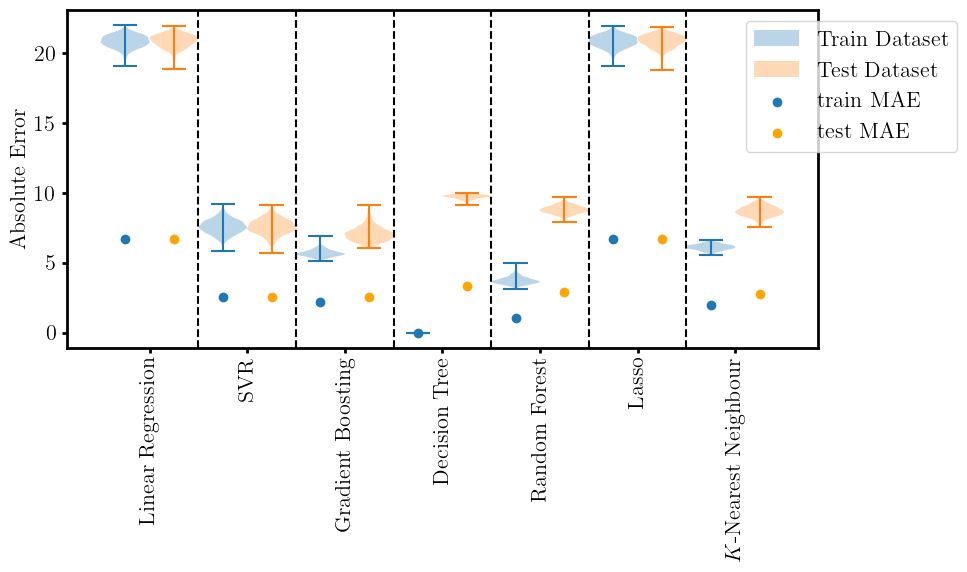}
    \caption{The figure highlights the variability between average and extreme errors across models. SVR stands out by reducing both averages and extremes, making it, for example, suitable for real-world applications. Overfitting is evident in models like the decision tree and random forest, where training errors are minimal, but test errors remain high. The visualization underscores the importance of analyzing extreme values, using methods like GEV or generalized Pareto distributions, to better understand worst-case scenarios.}
    \label{fig:multiple1}
\end{figure}

To further highlight the importance and the behavious of extreme errors, in Figure \ref{fig:multiple1} violin plots for the maximum of errors obtained with multiple models (so by training multiple model types) by using a blocking approach as described in Section \ref{sec:synthgen} are shown. 
The models trained are 
\begin{itemize}
    \item Linear Regression.
    \item Support Vector Regressor with the Radial Basis Function (RBF) kernel.
    \item Gradient Boosting Regressor with 100 estimators.
    \item Decision Tree Regressor.
    \item Random Forest Regressor.
    \item Lasso Regressor with the parameter $\alpha=1$.
    \item $K$-Nearest Neighbour Regressor with $K=3$.
\end{itemize}
All other parameters for the models have the standard values found in the scikit-learn Python library version 1.6.1 \cite{scikit-learn}. The fitted Extreme Value Distribution (EVD) parameters for all the models tested are reported in Table 
\ref{tab:evd_parameters}. The confidence intervales are calculated with a bootstrap approach with \(p=0.95\).
\begin{table}[htb]
\centering
\caption{Fitted extreme value distribution (EVD) parameters (\(\xi\) for shape, \(\mu\) for location, and \(\sigma\) for scale) for different models (with the blocking approach), along with their confidence intervals evaluated using a bootstrap approach. It is important to note that \(\mu\) and \(\sigma\) do not correspond to the mean and standard deviation of the data. For the decision tree regressor, the value from the training dataset are marked as N/A due to extreme overfitting, where all error values are zero, making it impossible to fit the data to a GEV family of distributions. Additionally, the table highlights how overfitting influences parameter values, as seen in the random forest regressor, where \(\xi\) differs significantly between the training (\(-0.087\)) and validation (\(-0.239\)) datasets, with substantial variation in confidence intervals. The confidence intervals are evaluated with \(p=0.95\).}
\label{tab:evd_parameters}
\adjustbox{max width=\textwidth}{
\begin{tabular}{@{}lcccc@{}}
\toprule
\textbf{Model}                    & \textbf{Dataset}            & \(\xi\) (Shape)                            & \(\mu\) (Location)                           & \(\sigma\) (Scale)                          \\ \midrule
\multirow{2}{*}{\textbf{Linear Regression}} 
& Training   & \(-0.376\) {[}-0.425, -0.343{]} & \(20.711\) {[}20.684, 20.736{]} & \(0.479\) {[}0.457, 0.502{]} \\
& Validation & \(-0.422\) {[}-0.458, -0.399{]} & \(20.734\) {[}20.710, 20.761{]} & \(0.508\) {[}0.486, 0.530{]} \\ \midrule

\multirow{2}{*}{\textbf{SVR}} 
& Training   & \(-0.271\) {[}-0.311, -0.239{]} & \(7.381\) {[}7.354, 7.412{]}   & \(0.522\) {[}0.502, 0.544{]} \\
& Validation & \(-0.294\) {[}-0.328, -0.267{]} & \(7.389\) {[}7.357, 7.418{]}   & \(0.550\) {[}0.529, 0.571{]} \\ \midrule

\multirow{2}{*}{\textbf{Gradient Boosting}} 
& Training   & \(-0.064\) {[}-0.126, -0.001{]} & \(5.578\) {[}5.564, 5.592{]}   & \(0.208\) {[}0.199, 0.218{]} \\
& Validation & \(-0.065\) {[}-0.102, -0.001{]} & \(6.858\) {[}6.831, 6.885{]}   & \(0.432\) {[}0.416, 0.450{]} \\ \midrule

\multirow{2}{*}{\textbf{Decision Tree}} 
& Training   & \textit{N/A}                 & \textit{N/A}                  & \textit{N/A}                \\
& Validation & \(-0.521\) {[}-0.546, -0.495{]} & \(9.691\) {[}9.682, 9.699{]}   & \(0.160\) {[}0.153, 0.167{]} \\ \midrule

\multirow{2}{*}{\textbf{Random Forest}} 
& Training   & \(-0.087\) {[}-0.128, -0.040{]} & \(3.606\) {[}3.595, 3.618{]}   & \(0.218\) {[}0.208, 0.227{]} \\
& Validation & \(-0.239\) {[}-0.282, -0.002{]} & \(8.690\) {[}8.665, 8.709{]}   & \(0.292\) {[}0.279, 0.305{]} \\ \midrule

\multirow{2}{*}{\textbf{Lasso}} 
& Training   & \(-0.390\) {[}-0.446, -0.353{]} & \(20.708\) {[}20.681, 20.738{]} & \(0.475\) {[}0.452, 0.497{]} \\
& Validation & \(-0.441\) {[}-0.478, -0.417{]} & \(20.731\) {[}20.703, 20.757{]} & \(0.503\) {[}0.481, 0.526{]} \\ \midrule

\multirow{2}{*}{\textbf{K-Nearest Neighbor}} 
& Training   & \(-0.308\) {[}-0.370, -0.003{]} & \(6.040\) {[}6.009, 6.054{]}   & \(0.210\) {[}0.202, 0.217{]} \\
& Validation & \(-0.277\) {[}-0.304, -0.250{]} & \(8.552\) {[}8.531, 8.573{]}   & \(0.377\) {[}0.361, 0.392{]} \\ \bottomrule
\end{tabular}
}
\end{table}

In Figure \ref{fig:multiple1} it can be clearly seen how the distance between maximum errors and averages vary from model to model. This adds additional information to models, namely, gives information about the distribution of errors and their extremes, showing how certain models not only give slightly lower averages, but increasingly lower extreme errors (such as, for example, SVR), making them more interesting for real-life applications. 

Overfitting manifests itself intriguingly within extreme value distributions. For instance, the outcomes derived from the decision tree regressor illustrate how the training dataset errors, even including extreme values, are essentially negligible, while the ones from the test dataset are not. Similarly, the random forest regressor also vividly demonstrates the occurrence of overfitting within extreme value distributions (the violin plot obtained from training data is made of clearly lower values than the one obtained from test data).


Since the importance of confidence intervals was highlighted previously, in Table \ref{tab:evd_parameters} the list of the fitted parameters for the different models and their confidence intervals evaluated with a bootstrap approach are listed. 

\subsection{EVT Applied to Real Data}
\label{sec:questionc}

To demonstrate the proposed methodology, we apply it to two real-world datasets: the Diabetes dataset \cite{diabetes_34} and the WHO Life Expectancy dataset \cite{life_expectancy_dataset}. 
The Diabetes dataset is a well-known benchmark for regression tasks in machine learning, containing physiological and medical measurements related to diabetes progression. The WHO Life Expectancy dataset provides a broader perspective, encompassing multiple health indicators and demographic factors from various countries.


\subsubsection{Diabetes Dataset}

The diabetes dataset \cite{diabetes_34} includes 10 input features representing various physiological and medical measurements, such as age, sex, BMI (body mass index), average blood pressure, and six blood serum measurements. The target variable is a quantitative measure of disease progression, making it a regression problem.
The dataset contains 442 samples. The results of the fit to the GEV family of distributions on the validation datasets are in Table \ref{tab:diab1} for linear regression. For reference the values obtained by fitting the generalised Pareto distribution are $\sigma = 5.4$, $\xi=-0.18$.

\begin{table}[htbp]
\centering
\caption{Parameters and Confidence Intervals for the Diabetes Dataset with the blocking approach  for linear regression.}
\label{tab:diab1}
\begin{tabular}{@{}lll@{}}
\toprule
\textbf{Parameter}        & \textbf{Value}  & \textbf{95\% Confidence Interval (CI)}      \\ \midrule
\(\xi\) (Shape)           & -0.324           & [-0.330,-0.320 ]                         \\
\(\mu\) (Location)        & 152.365         & [152.184, 152.562]                     \\
\(\sigma\) (Scale)        & 10.724          & [10.592, 10.860]                       \\ \bottomrule
\end{tabular}
\end{table}

\begin{figure}
    \centering
    \includegraphics[width=1.0\linewidth]{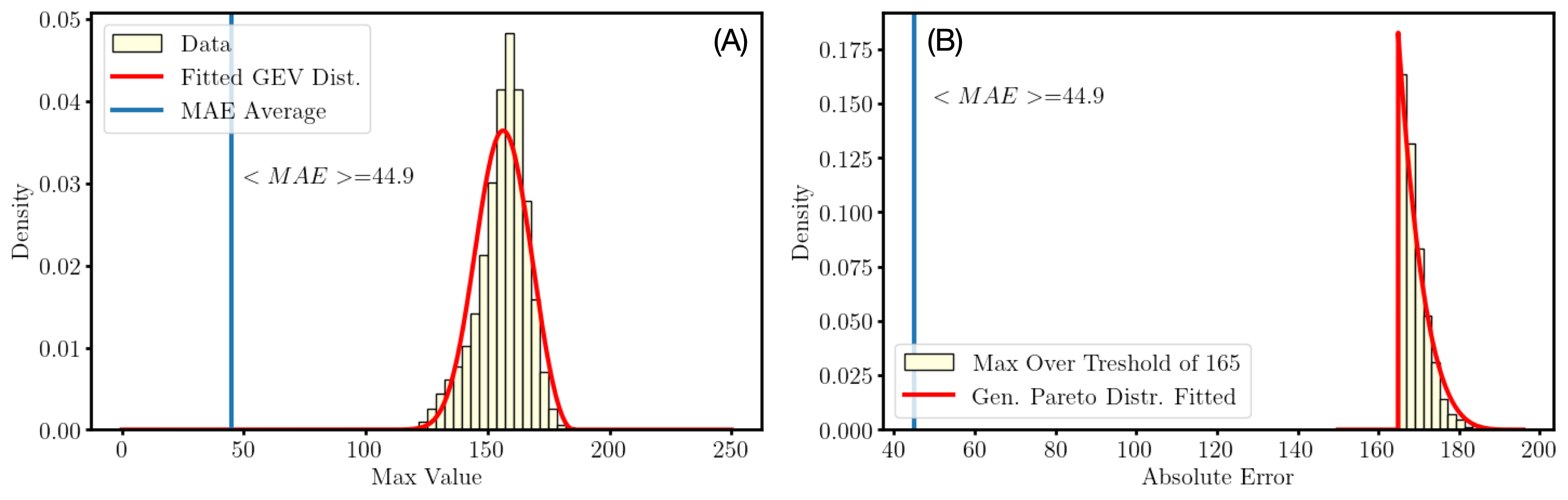}
    \caption{Results of extreme value analysis on the Diabetes dataset. (A) Blocking approach with 80\%/20\% Monte Carlo cross-validation, showing the GEV fit (red) and MAE (blue). (B) Generalized Pareto fit for errors exceeding a threshold of 165. }
    \label{fig:diab1}
\end{figure}

In Figure \ref{fig:diab1} the results for the blocking approach (with a 80\%/20\% based monte carlo CV) (panel A) and for values exceeding a treshold of 165 (panel B)  are shown. One can now estimate that linear regression, applied to this dataset, will give an error smaller than 177.2 in 95\% of cases (when using the blocking approach). It is evident that $<\textbf{MAE}>=44.9$ provides an overly optimistic evaluation of the performance of the model.

\subsubsection{WHO Dataset}
The WHO Life Expectancy dataset \cite{life_expectancy_dataset} provides a comprehensive view of global health indicators and their influence on life expectancy across 193 countries from 2000 to 2015. 
This dataset has undergone the following preprocessing procedures.
\begin{itemize}
    \item Non-numerical features have been removed.
    \item The features \textit{Year} and \textit{Population} has been removed. As life expectancy varies over time and nations evolve, using the year as an input feature may result in models that improperly prioritise the year over more salient features (since the year might be interpreted as a proxy for the important features). This issue similarly applies to the \textit{population} feature.
    \item All numerical features are normalised to have a mean of zero and a standard deviation of one.
    \item All observations with a \textit{nan} (empty value) have been removed. Thus, the final dataset has 1853 observations and 18 features.
\end{itemize}

The results of the fit to the GEV family of distributions on the validation datasets are in Table \ref{tab:life1} for linear regression. 
Linear regression predictions will have an error not larger than 25.6 years in 95\% of the cases (when using the blocking approach). Note that the average MAE is 2.8 years, which would make an observer think that the performance of the model is exceptional, while this is a clearly an overly optimistic (and wrong) assessment.
\begin{figure}
    \centering
    \includegraphics[width=1\linewidth]{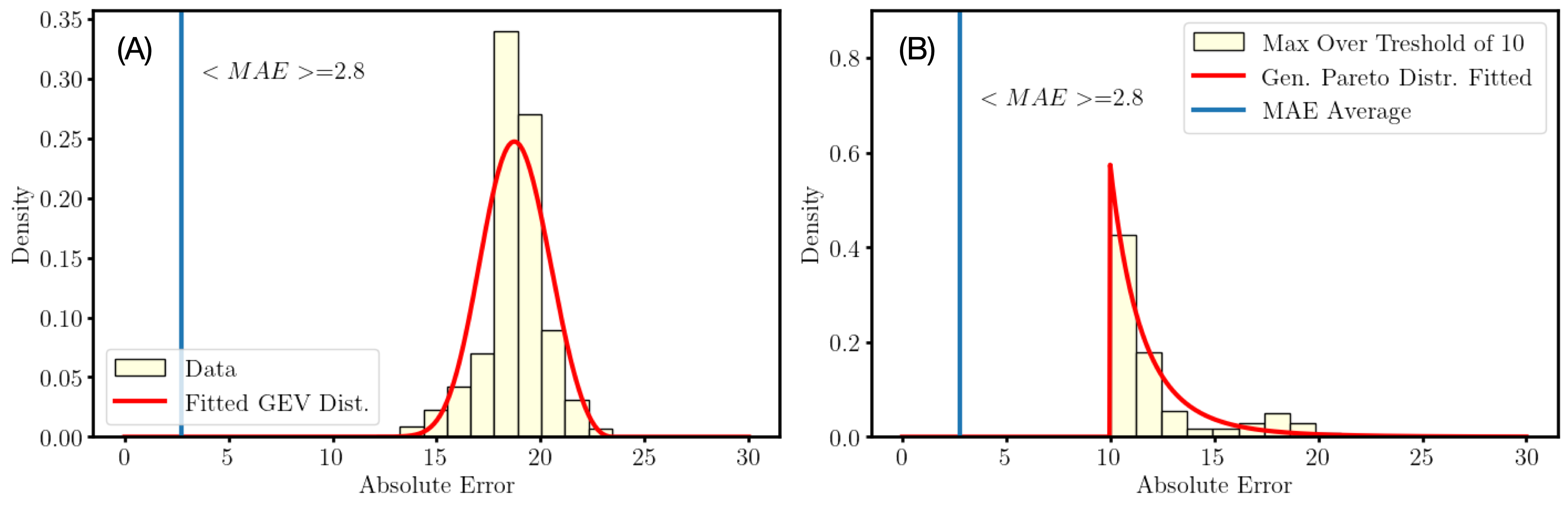}
    \caption{Results of extreme value analysis on the WHO dataset. (A) Blocking approach with 80\%/20\% Monte Carlo cross-validation, showing the GEV fit (red) and MAE (blue). (B) Generalized Pareto fit for errors exceeding a threshold of 10.}
    \label{fig:life1}
\end{figure}

\begin{table}[htbp]
\centering
\caption{Parameters and Confidence Intervals for the WHO Dataset with the blocking approach  for linear regression.}
\label{tab:life1}
\begin{tabular}{@{}lll@{}}
\toprule
\textbf{Parameter}        & \textbf{Value}  & \textbf{95\% Confidence Interval (CI)}      \\ \midrule
\(\xi\) (Shape)           & -0.296           & [-0.308,-0.287 ]                         \\
\(\mu\) (Location)        & 18.217          & [18.181, 18.261]                       \\
\(\sigma\) (Scale)        & 1.559           & [1.525, 1.591]                         \\ \bottomrule
\end{tabular}
\end{table}


\section{Discussion}
\label{sec:con}

This paper proposes the application of Extreme Value Theory to machine learning. With EVT, it is now possible to quantify the probabilities of extreme errors, providing a statistically robust approach to evaluate and compare model performance under worst-case scenarios.
The proposed methodology was validated on both synthetic and real-world datasets, demonstrating its versatility and effectiveness. Through synthetic experiments, it is showcased how EVT can model maximum errors and threshold exceedances. The real-world applications to the Diabetes dataset and the WHO Life Expectancy dataset further highlight the importance of understanding extreme errors, especially in high-stakes fields like medicine and public health.

The ability to quantify and predict extreme errors has profound implications for high-stakes domains, including medical diagnostics, financial modeling, and public policy, where understanding the risks of worst-case scenarios is critical. In addition, in fields like Physics or Chemistry it will now be possible to estimate the maximum error of a model when applied to new data and thus it will be possible to assess \textit{model error}, a concept akin to the measurement errors of instruments, transforming machine learning algorithms in real \textbf{instruments} for science.


Future work will focus on extending this approach to classification problems, and investigating the interplay between dataset size and the reliability of EVT-based predictions. Furthermore, future research will aim to refine the EVT parameter estimation methods for improved robustness in the presence of small data sets or high-dimensional data.

While the proposed application of Extreme Value Theory (EVT) to machine learning model validation introduces a novel framework for assessing extreme errors, there are limitations that needs to be discussed. Firstly, EVT assumes that the observed data follow a stationary distribution, meaning that the statistical properties of errors remain constant over time. However, in real-world machine learning applications, model errors may evolve due to changes in data distribution (e.g., concept drift). This means that as with models, parameters and distributions must be checked and possibly refitted when new data appear. Secondly, if one wants to avoid the blocking procedure, the accuracy of EVT models depends on an appropriate choice of the threshold \( u \) when applying the generalized Pareto distribution (GPD) approach (see Section \ref{sec:choosingu}), something that is not easyy to determine. Thrdily, EVT is theoretically derived in the limit of large sample sizes, but practical applications often involve limited data. When datasets are small, the estimation of the probabilities of extreme events can become less reliable, leading to wider confidence intervals and increased uncertainty in the predictions. Furthermore, the proposed methodology requires repeated model training and validation using Monte Carlo cross-validation, which can be computationally expensive. More research is needed to determine suitable EVT-based methodologies. Furthermore, more work is needed to develop user-friendly tools for integrating EVT-based risk assessment into practical machine learning workflows.

\section{Methods}
\label{sec:methods}

\subsection{Extreme Value Theory (EVT) Short Introduction}
\label{sec:evt}

Consider an array of random variables $X_1$, $X_2$, $\cdots$ and denote their maximum over an ``$n$-observation'' period with
\begin{equation}
    M_n = \max \{  X_1, \cdots, X_n\}
\end{equation}
Grouping random variables in $n$-oberservations period is called in the EVT jargon \textit{blocking}.
In general the behaviour of the $X_i$ (in other words their distribution) is unknown, making exact evaluations of the distribution of $M_n$ impossible. Extreme Value Theory (EVT) tells us that, under specific assumptions, the behaviour of $M_n$ (to be precise when rescaled, more on that later) for $n\rightarrow \infty$ leads to a family of known distributions \cite{coles_introduction_2001, de_haan_extreme_2006, de_haan_extreme_2006}. For simplicity of notation, $x_i$ will indicate the realisations of the random variables $X_i$ (in other words, their measured values). One key assumption to make is that the distributions of $X_i$ do not depend on $i$. For example, if the index $i$ indicates the time of measurement, the assumption is that the distribution of $X_i$ is constant and does not depend on when they are measured. For example, if $X_i$ indicates the amount of rain in a specific location at a specific time, one has to be careful. If $n$ is too large, the measurement window may encompass winter and summer. The $X_i$ measured in winter and those measured in summer will naturally have different distributions \cite{coles_introduction_2001}. In this case EVT results are not valid and cannot be used. The main theorems of EVT is the following.
\begin{theorem}
\label{theorem:gev1}
    If there exist a sequence of constants $a_n>0$ and $b_n$ such that
    \begin{equation}
        P\left[(M_n-b_n)/a_n\leq z\right] \rightarrow G(z) \ \ \textrm{as} \ n\rightarrow \infty
    \end{equation}
    for a non-degenerate distribution $G$, then $G$ is a member of the Generalised Extreme Value (GEV) family
    \begin{equation}
        G(z)=\exp \left\{
            -\left[
                1+\xi \left(  \frac{z-\mu}{\sigma} \right)
            \right]^{-1/\xi}
        \right\}
    \end{equation}  
    defined on $\{z: 1+\xi(z-\mu)/\sigma>0\}$, where $-\infty<\mu<\infty$, $\sigma>0$ and $-\infty < \xi < \infty$.
\end{theorem}
For all practical purposes, this theorem will be used for \textit{large} values of $n$ as an approximation for the limit $n\rightarrow \infty$. When fitting data to this family of distributions, the determination of the sequences $a_n$ and $b_n$ is not relevant, as we are only interested in the values of the constants $\xi$, $\mu$ and $\sigma$ for the limit of large $n$. Since an analytical formula for $G$ is known, the maximum likelihood method \cite{coles_introduction_2001} can be used to determine the parameters of the GEV distribution that fit the data. 

In addition to \textit{blocking} there is another approach that is possible, using the \textit{threshold method}. In some cases it is better to avoid altogether the procedure of blocking. It is possible to define as extreme events (the one we are interested in studying) those where $X_i$ exceeds some high threshold $u$. 
Generally speaking we are interested in
\begin{equation}
    P(X>u+x | X>u)
\end{equation}
with $u,x \in \mathbb{R}$ and $u,x>0$.
In this case the following theorem can be proved \cite{coles_introduction_2001}.
\begin{theorem}
Let $X_1$, $X_2$, ... be a sequence of independent random variables with common distribution function $F$.   Suppose that $F$ satisfy Theorem \ref{theorem:gev1}. Then for large enough $u$, the distribution function of $(X-u)$, conditional on $X>u$ is given by approximately
\begin{equation}
\label{eq:gpd}
    H(z) = \begin{cases} 
        1 - \left( 1 + \frac{\xi(x - \mu)}{\sigma} \right)^{-1/\xi} & \text{for } \xi \neq 0 \\ 
        1 - \exp \left( -\frac{(x - \mu)}{\sigma} \right) & \text{for } \xi = 0 
    \end{cases}
\end{equation}
for $x \geq \mu$ when $\xi \geq 0$, and $\mu \leq x \leq \mu-\sigma/\xi$ when $\xi<0$.
\end{theorem}
The family of distributions defined in \ref{eq:gpd} is called the \textbf{Generalised Pareto Distribution} (GPD) family.

An important point is that the parameters of the GPD are uniquely determined by those of the associated GEV distribution of block maxima. In particular, the parameter $\xi$ mentioned in Equation (\ref{eq:gpd}) corresponds to the corresponding parameter in the associated GEV distribution. 



\subsection{Fitting Data to the GEV and GPD distribution families}

Once data are available (the $M_n$ or values exceeding a threshold $u$), it is possible to fit their distributions with the maximum likelihood approach, as described in detail in \cite{coles_introduction_2001}. It is of fundamental importance to estimate confidence intervals for the parameters $\xi$, $\mu$ and $\sigma$. This is especially relevant for the $\xi$ parameter. In fact, if $\xi$ changes sign, the type of distribution changes. One must be particularly careful in the case that the confidence intervals for $\xi$ cover positive and negative values for $\xi$.
if values of $\xi$ in the confidence intervals fall within small windows around zero, one should use the limit for $\xi\rightarrow 0$ of Equation (\ref{theorem:gev1}) (called the Gumbel limit of the GEV distribution). In general, first one should estimate the parameters with the maximum likelihood estimation. If the confidence intervals of the shape parameter $\xi$ include zero, one should test the alternate hypothesis that $\xi$ may be zero and use the Gumbel limit of the GEV distribution \cite{coles_introduction_2001}. There is no real agreement on how to properly treat such cases \cite{coles_introduction_2001}. 

\subsection{Choosing the Treshold $u$}
\label{sec:choosingu}

The choice of the threshold \( u \) is typically guided by the specific use case being analysed. However, it is also important to select \( u \) within a range in which the estimated parameters of the GEV family remain relatively stable. Stability in these parameters suggests that the extreme value model is capturing the tail behaviour consistently, reducing the risk of biased or unreliable estimates of extreme error behaviour.

The \textit{parameter stability plot} is a widely used method to determine the appropriate threshold \( u \) when applying the Generalised Pareto Distribution (GPD) to model exceedances. In general, this approach works quite well for the Frechet and Gumbell families of GEV distributions (arising with the blocking approach) and is less easy to use for the Weibull one. Let us describe the main idea, that is, to evaluate the stability of the estimated shape parameter \( \xi \) over a range of threshold values. The optimal threshold \( u \) is chosen where the estimates of \( \xi \) remain approximately constant.
To construct the parameter stability plot, one proceed as follows:

\begin{enumerate}
    \item Select a range of candidate threshold values \( u_1, u_2, \dots, u_k \).
    \item For each \( u_i \), consider the subset of the dataset consisting of exceedances over \( u_i \), denoted as
    \begin{equation}
        X^{(i)} = \{ X_j - u_i \mid X_j > u_i \}.
    \end{equation}
    \item Fit a Generalized Pareto Distribution (GPD) to \( X^{(i)} -u\) and estimate the shape parameter \( \xi(u_i) \) and scale parameter \( \sigma(u_i) \).
    \item Plot \( \xi(u_i) \) against the threshold \( u_i \).
    \item Identify the range of \( u \) where \( \xi(u) \) remains approximately constant and select the smallest threshold within this stable region.
\end{enumerate}
If $\xi(u)$ fluctuates significantly with $u$, it suggests that lower thresholds include too many non-extreme values. Conversely, if $\xi(u)$ stabilizes over a range of $u$, this indicates an appropriate threshold selection. A sudden increase in $\xi$ at higher thresholds may be due to too few exceedances, leading to unreliable estimates.


The \textit{parameter stability plot} is not the only method to choose $u$. Other methods (not discussed here) are the Mean Residual Life (MRL) Plot, automatic threshold selection via statistical tests between others.

\subsection{Application of EVT to Machine Learning}
\label{sec:evtml}

Consider a scenario in which a particular model is employed to forecast a physical quantity through a regression task on a specified dataset $D$. Currently, establishing a statistically reliable estimate for the maximum error the model might incur with novel data is not possible. To address this deficiency, one should assess the distribution of all extreme errors as described previously. Using the fitted distribution, one can select, for instance, the 99\% or 95\% quantile as the experimental error \textit{of the model} depending on the confidence level required. This methodology enables the implementation of machine learning models for regression tasks that involve scientific data, while simultaneously offering a statistically robust estimate of the experimental error \textit{of the model}.

\begin{figure}
    \centering
    \includegraphics[width=0.9\linewidth]{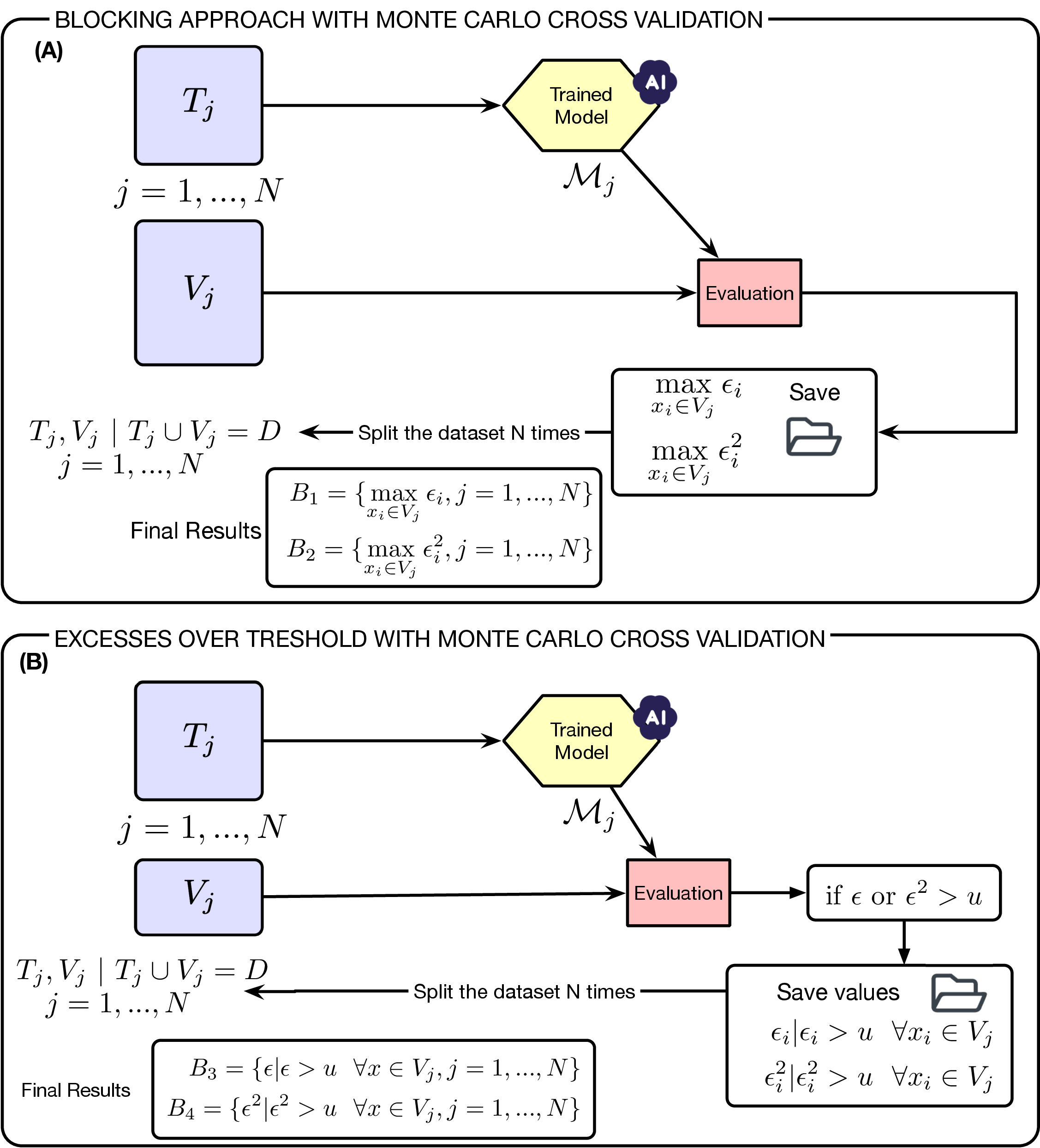}
    \caption{Panel (A) describes the application of EVT in conjunction with monte-carlo cross validation with the blocking approach. Panel (B) describes the apllicatio of EVT in conjunction with monte-carlo cross validation with errors exceeding a given treshold $u$.}
    \label{fig:evt_ml}
\end{figure}

The exact process can be performed in conjunction with Monte Carlo cross-validation \cite{michelucci2024fundamental} and is depicted in detail in Figure \ref{fig:evt_ml}. The process works according to the following steps.
\begin{enumerate}
    \item The dataset $D$ is split in two portions $T_j$ and $V_j$ according to some proportions, such that $T_j\cup V_j =D$.
    \item The model ${\cal M}$ is trained on the training portion of the dataset $T_j$. The trained model will be indicated with ${\cal M}_j$
    \item The model is evaluated on the validation portion of the dataset $V_j$. 
    \item With the trained model ${\cal M}_j$ it is now possible to evaluate errors $\epsilon_i \equiv {\cal M}_j(x_i)-y_i$ $\forall x_j\in V_j$ and $\epsilon_i^2 \equiv ({\cal M}_j(x_i)-y_i)^2$ $\forall x_j\in V_j$ (where $y_i$ indicates the expected value for observation $x_j$).
    \item Now, it is possible for each split to save the maximum of errors (panel (A) in Figure \ref{fig:evt_ml}) or to save all errors values that exceed a given threshold (panel (B) in Figure \ref{fig:evt_ml}).
    \item This process is performed $N$ times (step 1-5).
    \item Finally, the datasets saved in the previous step must be fitted to the GEV or GPD families, thus obtaining the distributions for the maxima as discussed.
\end{enumerate}

It should be emphasised that this approach introduces a revolutionary capability in machine learning: the ability to quantify the probability that a model class (for example neural networks or random forest) will make, as largest error or one exceeding a given threshold $u$. By applying Extreme Value Theory (EVT) to machine learning, we move beyond traditional measures of model performance like average error or variance, and we move into the realm of extreme events (the largest error), those critical occurrences that can have a large impact. This advancement opens the way for a deeper understanding of worst-case scenarios, which are often overlooked in conventional analyses.
The implications are particularly profound in high-stakes applications where decisions have significant consequences. For example, in medicine, the ability to predict the likelihood of large errors could transform risk management, allowing physicians to better evaluate the safety and reliability of AI-driven diagnostics or treatment recommendations. This could lead to more informed judgments about what levels of risk are acceptable, ensuring that critical decisions are grounded in a robust understanding of potential failures.
Furthermore, this approach can help establish new standards for model evaluation, helping in the development of machine learning systems designed for high-reliability environments. By shedding light on the tail-end behaviour of errors, this methodology not only enhances our capacity to anticipate and mitigate risks, but also fosters trust in AI systems, which is essential for their adoption in domains where human lives, safety, and well-being are at stake.

\bibliographystyle{unsrt}
\bibliography{biblio.bib}

\end{document}